# Modeling Information Incorporation in Markets, with Application to Detecting and Explaining Events

**David M. Pennock*, Sandip Debnath*,†, Eric J. Glover* and C. Lee Giles*,†,‡**
*NEC Research Institute
4 Independence Way
Princeton, NJ
08540 USA
†Department of Computer Science and Engineering
‡School of Information Sciences and Technology
Pennsylvania State University
University Park, PA 16801 USA
{dpennock,debnath,compuman,giles}@research.nj.nec.com

## Abstract

We develop a model of how information flows into a market, and derive algorithms for automatically detecting and explaining relevant events. We analyze data from twenty-two "political stock markets" (i.e., betting markets on political outcomes) on the Iowa Electronic Market (IEM). We prove that, under certain efficiency assumptions, prices in such betting markets will on average approach the correct outcomes over time, and show that IEM data conforms closely to the theory. We present a simple model of a betting market where information is revealed over time, and show a qualitative correspondence between the model and real market data. We also present an algorithm for automatically detecting significant events and generating semantic explanations of their origin. The algorithm operates by discovering significant changes in vocabulary on online news sources (using expected entropy loss) that align with major price spikes in related betting markets.

## 1 INTRODUCTION

In markets where items sold have uncertain value, an equilibrium price can be viewed as a summary statistic reflecting the knowledge of all market players about the item's value. In fact, under common efficiency assumptions, price is a *sufficient* statistic for the value of an item, given all evidence known to all participants in the market. In this sense, markets are very effective mechanisms for aggregating information that is spread across a population, summarizing that information concisely in the form of (often publicly available) prices.

Gambles are particularly extreme examples of uncertain-value items.[1] A gamble (also called a *security*) pays an amount contingent on some future outcome. For example, the gamble "$1 if it rains tomorrow" pays $1 if it rains tomorrow, and $0 if it does not rain. If an agent purchases (one unit of) this gamble for $0.3, then the agent wins $1-$0.3=$0.7 if it rains, and loses the $0.3 otherwise. Tomorrow, the value of the gamble will be certain (exactly $1 or exactly $0). But the value of the gamble today depends on the probability of rain tomorrow. In fact, assuming the buyer is risk neutral, the value to the buyer is exactly his or her subjective probability of rain, since the buyer's expected value is E[$1 if RAIN] = Pr(RAIN)·$1 + Pr(NO-RAIN)·$0 = $Pr(RAIN).

Most hypotheses about the efficiency of markets assume that information is incorporated into market prices virtually instantaneously, as soon as it becomes available to any trader. Informally, the reasoning is that, if some trader has superior information that allows him or her to obtain an expected profit at the current price, then he or she will take advantage of the opportunity by appropriately buying or selling, thereby driving prices toward the correct value given the new information.

In this paper, we develop a theory and model of information incorporation in markets, and compare our theoretical predictions with real data from twenty-two political "stock markets" (betting markets on political elections) on the Iowa Electronic Market (IEM).[2] We present an application where markets can be used as detection devices for significant information events, and parallel text streams (e.g., newsgroups or news media) can be mined to give semantic explanations of the events.

In Section 2, we give necessary background informa-tion and introduce notation. In Section 3, we present results of a large-scale analysis of the IEM. In Section 4, we prove that, under certain efficiency assumptions, and without an explicit model of evidence or information, the prices in betting markets will over time converge (in expectation) toward the eventual outcome. For example, the price of "$1 if RAIN" will tend to rise over time in worlds where RAIN is true, as compared to worlds where RAIN is false, according to a very simple relationship that appears to hold on IEM. In Section 5, we give a model of information release and subsequent incorporation in a betting market, and show a strong qualitative correspondence with IEM price dynamics. In Section 6, we present our algorithm for detecting and explaining information events. The algorithm looks at changes in vocabulary before and after significant market swings by ranking words and phrases according to expected entropy loss. We show that in three cases, the algorithm gives appropriate and meaningful semantic explanations for large price changes.

[1] In fact, gambles are often set up purposely to maximize uncertainty (entropy), for example by setting lines in a sporting event to make the outcome a fifty-fifty proposition.

[2] http://www.biz.uiowa.edu/iem/

# 2 INFORMATION INCORPORATION IN BETTING MARKETS

## 2.1 BACKGROUND

It is clear that markets often react quickly to the release of new and relevant information. For example, on March 11, 2002, the stock of Cepheid rose 34% when the company announced it was developing biological hazard detection devices for mail systems. Perhaps the relationship between price and information is no more clear than in betting markets. The current price of a gamble is precisely related to the probabilities of the possible payoffs of the gamble, and any information available that acts as evidence for the possible outcomes should affect the price appropriately according to the rules of Bayesian updates.

The economic theory of *rational expectations* (RE) accounts for information incorporation in markets. RE theory posits that prices reflect the sum total of all information available to all market participants (Grossman, 1981; Lucas, 1972). Even when some agents have exclusive access to inside information, prices equilibrate exactly as if everyone had access to all information. The procedural explanation is that prices reveal to the ignorant agents any initially private information; that is, agents learn by observing prices.

Plott et al. (1997) investigate, in a laboratory setting, whether parimutuel markets (the type employed at horse races) are able to aggregate information, as postulated by RE theory. In one set of experiments, each subject was given inside knowledge that a subset of horses would definitely *not* win. Although all subjects were uncertain as to the outcome, their collective information was enough to identify the winning horse with certainty. Information aggregation did occur, and RE-based predictions fit the data well.

Plott and Sunder (1982, 1988) and Forsythe and Lundholm (1990) conducted laboratory experiments to test the reasonableness of the RE assumption in the context of a securities market (essentially a betting market as described in the introduction). In many cases, even when information was distributed asymmetrically accross participants (for example, certain traders were given "inside" knowledge or evidence pointing toward particular outcomes), the equilibrium reached reflected the combination of all information, as predicted by RE theory.

Beyond the controlled setting of the laboratory, empiricists have analyzed the accuracy of implied probability assessments given by public markets. Perhaps the most direct tests involve horse race betting markets. Several studies demonstrate that odds on horses correlate well with the actual frequencies of victory (Thaler & Ziemba, 1988). Other sports betting markets, like the National Basketball Association point spread market, provide very accurate forecasts of likely game outcomes (Gandar, Dare, Brown, & Zuber, 1998). Financial options markets (in many ways equivalent to betting markets) yield accurate probability distributions over the future prices of their underlying stocks (Sherrick, Garcia, & Tirupattur, 1996).

The Iowa Electronic Market (IEM) supports trading in securities tied to the outcome of political and financial events. Their 1988 market, open only to University of Iowa students and employees, offered securities that paid off proportionally to the percentage of votes received by various candidates in that year's US Presidential election. The final prices matched Bush's final percent margin of victory more closely than any of the six major polls (Forsythe, Nelson, Neumann, & Wright, 1992). Since opening to the public, subsequent US Presidential election markets have attracted wide participation and following. Other election markets have now opened in Canada[3] and Austria.[4]

We use the *logarithmic score* to measure accuracy and information incorporation in IEM. The logarithmic score is a *proper scoring rule* (Winkler & Murphy, 1968), and is an accepted method of evaluating probability assessments. When experts are rewarded according to a proper score, they can maximize their expected return by reporting their probabilities truth-fully. Additionally, more accurate experts can expect to earn a higher average score than less competent experts. Suppose an expert reports probabilities $p_1, p_2, \ldots, p_k$ for $k$ mutually exclusive and exhaustive alternatives. Let $w_i = 1$ if and only if the $i$th event occurs, and $w_i = 0$ otherwise. Then the expert's score for the current event is $\ln\left(\sum_{i=1}^{k} w_i p_i\right)$. Higher scores indicate more accurate forecasts, with 0 the maximum and negative infinity the minimum. The "expert assessments" given by the market are taken to be the (normalized) prices of the candidates.

[3] http://esm.ubc.ca/

[4] http://ebweb.tuwien.ac.at/apsm/

Note that under the logarithmic scoring rule, an expert's expected score equals the entropy of his or her probability distribution. Stated another way, the negative of the logarithmic score gives the amount that the expert is "surprised" by the actual outcome. So the logarithmic score applied to IEM is both a measure of forecast accuracy and an information-theoretic measure of the amount that the market is surprised when the winners of the elections are finally determined.

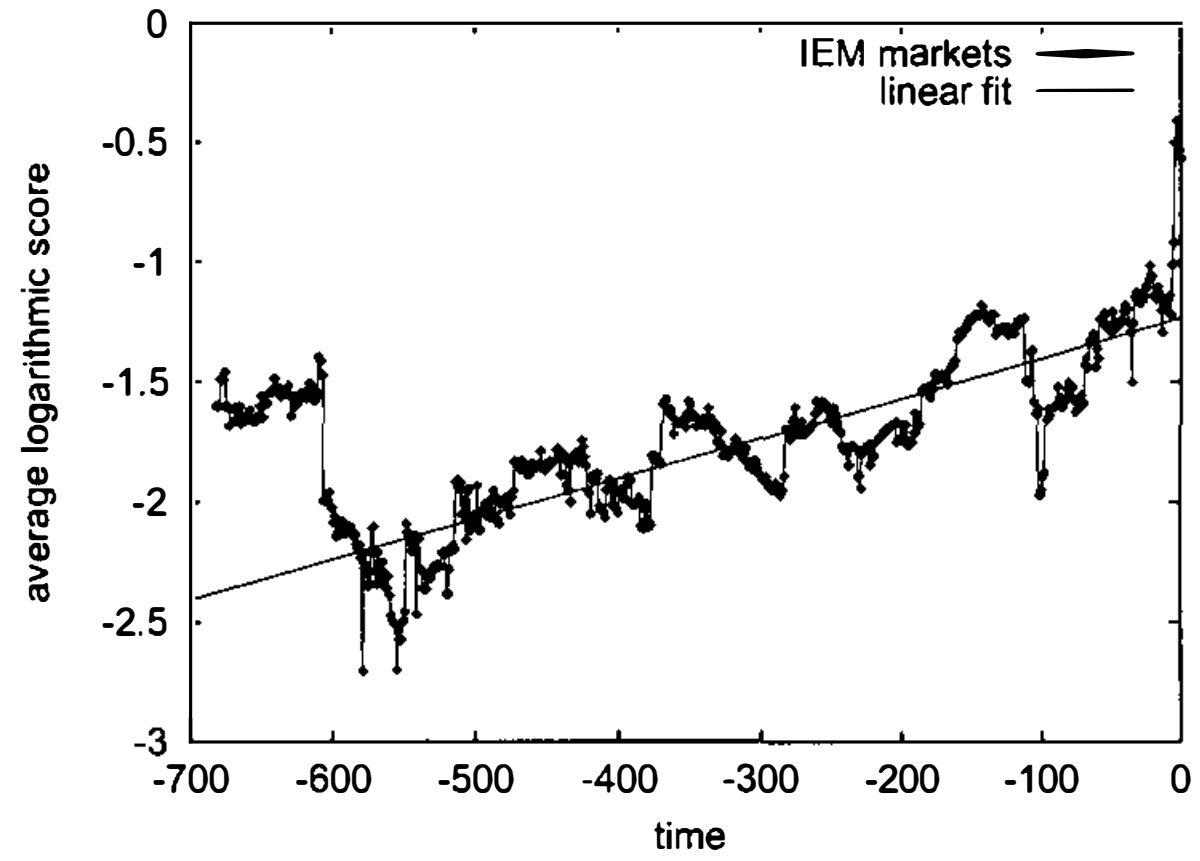

Figure 1: Average logarithmic score of twenty-two election markets on IEM. Higher (less negative) scores reflect increasing accuracy.

## 2.2 NOTATION

If the probability of event $E$ is $\Pr(E)$, then the *likelihood* of event $E$ is $\mathcal{L}(E) = \Pr(E)/\Pr(\bar{E})$ and the *log-likelihood* of $E$ is $\mathcal{LL}(E) = \ln \mathcal{L}(E) = \ln(\Pr(E)/\Pr(\bar{E}))$. We denote a gamble paying off \$1 if and only if event $E$ occurs as $\langle E \rangle$. Let the price of $\langle E \rangle$ at time $t$ be $p_t$. In analogy to the definitions of likelihood and log-likelihood, define the *likelihood price* as $l_t = p_t/(1-p_t)$ and the *log-likelihood price* as $ll_t = \ln l_t = \ln p_t/(1-p_t)$. So, for example, $\Pr(ll_t = b | ll_{t-1} = a)$ denotes the probability that the log-likelihood price equals $b$ at time $t$, given that the log-likelihood price equals $a$ at time $t-1$. Similarly, the probability that event $E$ occurs given that the likelihood price at time $t$ is $a$ is written as $\Pr(E | l_t = a)$, etc.

# 3 EXAMINING THE IOWA ELECTRONIC MARKET

We collected daily prices from twenty-two political election markets on IEM. Markets include the 2000 US Presidential election, the 2000 NY Senate election, and other elections in the US and around the world. In the NY Senate election, for example, traders could buy or sell shares of "\$1 if Giuliani wins", "\$1 if H. Clinton wins", "\$1 if Lazio wins", "\$1 if another republican wins", "\$1 if another democrat wins", and "\$1 if an independent candidate wins". On IEM, candidates can be sold by first buying the bundle of all candidates from the "bank" for \$1 (the bank is guaranteed to exactly break even with this transaction), then selling particular candidates back on the open market.

Figure 1 shows the average logarithmic score over time for all twenty-two markets. Time zero is aligned with the last trading day in every market (on or near the corresponding election day), and time $-i$ corresponds with $i$ days before the last trading day for every market. The plot at point $-i$ is an average over all markets that lasted at least $i$ days (fewer and fewer markets are represented toward the left end of the graph). The logarithmic score trends upward over time, indicating that the market's probability assessments are improving over time. Put another way, prices tend to move in the direction of the winning candidate. Morover, the increase in logarithmic score is roughly linear over a large portion of the graph, with a rapid rise evident just before time zero. It is important to note that recognizing this trend is only possible *after* the election outcomes are known, since the logarithmic score computation depends on the actual outcome. The fact that the logarthmic score increases over time does not mean that price movements are predictable *before* the election outcome is known (in fact, if the market is efficient, prices are not predictable).

The increase in logarithmic score can be attributed to the incorporation of evidential information into the market as it becomes available to traders. The linear increase over much of the graph can be thought of as a manifestion of a roughly constant flow of information on average into the markets. We discuss interpretations of the increase in logarithmic score in terms of our theory and model in Sections 4 and 5, respectively.

# 4 THEORY OF INFORMATION INCORPORATION

We first make some efficiency assumptions, then prove consequences for price dynamics in betting markets like IEM. We begin with a basic assumption about the accuracy of market prices as probability assessments.

**Assumption 1 (Forecast accuracy)** *Let $p_t$ be the price of $\langle E \rangle$ at time t. Then*

$$\Pr(E|p_t, p_{t-1}, p_{t-2}, \ldots, p_0) = p_t.$$

Assumption 1 says that the market price today is an accurate assessment of the probability of $E$, independent of any past prices. To see why Assumption 1 is reasonable, imagine for a moment that a betting market existed where Assumption 1 did not hold. Then a trader whose probability assessments are more accurate than the market's could earn a consistent profit (in expectation). But that trader's actions would act precisely to "correct" the market assessment, driving prices so as to make Assumption 1 true. RE theory essentially assumes that no individual trader has more accurate probability assessments than the market.

**Theorem 1** *Assumption 1 implies the following consequences:*

1.

$$E[p_t|p_{t-1} = a] = a,$$

*where $E[\cdot]$ is the expectation operator, not to be confused with the event $E$. That is, prior to knowing the outcome of $E$, the expected price at time $t$ equals the price at time $t-1$ (i.e., the apriori expected change in price is zero).*

2.

$$\frac{\Pr(ll_t = a + \epsilon|E, ll_{t-1} = a)}{\Pr(ll_t = a + \epsilon|\bar{E}, ll_{t-1} = a)} = e^\epsilon$$

*That is, the log-likelihood price is $e^\epsilon$ times as likely to go up by $\epsilon$ in worlds where $E$ is true as it is to go up by $\epsilon$ in worlds where $E$ is false.*

3.

$$\Pr(ll_t = a + \epsilon|E, ll_{t-1} = a) = \frac{e^a + 1}{e^a + e^{-\epsilon}} \Pr(ll_t = a + \epsilon|ll_{t-1} = a)$$

*That is, the log-likelihood price is $(e^a + 1)/(e^a + e^{-\epsilon})$ times as likely to go up from $a$ to $a+\epsilon$ in worlds where $E$ is true as it is to go up from $a$ to $a+\epsilon$ in worlds where the state of $E$ is unknown.*

4.

$$E[p_t|E, p_{t-1} = a] = a + \frac{\mathrm{Var}(p_t|p_{t-1} = a)}{a}$$

*That is, the expected price at time $t$ in worlds where $E$ is true is greater than the price at time $t-1$ by an amount proportional to the variance of price.*

**Proof.** We begin by proving item #1. By Assumption 1 and Bayes' rule,

$$\begin{aligned} a &= \Pr(E|p_{t-1} = a) \\ &= \int_0^1 \Pr(E|p_t, p_{t-1} = a) \Pr(p_t|p_{t-1} = a) dp_t \\ &= \int_0^1 p_t \Pr(p_t|p_{t-1} = a) dp_t \\ &= E[p_t|p_{t-1} = a] \end{aligned}$$

□

Next we prove item #2. Applying Bayes' rule using log-likelihood notation, we get:

$$\mathcal{LL}(E|ll_t = a+\epsilon, ll_{t-1} = a) = \ln \frac{\Pr(ll_t = a + \epsilon|E, ll_{t-1} = a)}{\Pr(ll_t = a + \epsilon|\bar{E}, ll_{t-1} = a)} + \mathcal{LL}(E|ll_{t-1} = a)$$

Applying Assumption 1, we get:

$$a + \epsilon = \ln \frac{\Pr(ll_t = a + \epsilon|E, ll_{t-1} = a)}{\Pr(ll_t = a + \epsilon|\bar{E}, ll_{t-1} = a)} + a$$

$$\frac{\Pr(ll_t = a + \epsilon|E, ll_{t-1} = a)}{\Pr(ll_t = a + \epsilon|\bar{E}, ll_{t-1} = a)} = e^{a+\epsilon-a} = e^\epsilon$$

□

Next we prove item #3. Summing over possible worlds ($E$ and $\bar{E}$), we get:

$$\begin{aligned} &\Pr(ll_t = a + \epsilon|ll_{t-1} = a) = \\ &\quad \Pr(ll_t = a + \epsilon|E, ll_{t-1} = a) \Pr(E|ll_{t-1} = a) + \\ &\quad \Pr(ll_t = a + \epsilon|\bar{E}, ll_{t-1} = a) \Pr(\bar{E}|ll_{t-1} = a) \end{aligned}$$

Applying Assumption 1, we get:

$$\begin{aligned} &\Pr(ll_t = a + \epsilon|ll_{t-1} = a) = \\ &\quad \Pr(ll_t = a + \epsilon|E, ll_{t-1} = a)\frac{e^a}{1+e^a} + \\ &\quad \Pr(ll_t = a + \epsilon|\bar{E}, ll_{t-1} = a)\frac{1}{1+e^a} \end{aligned}$$

Dividing both sides by $\Pr(ll_t = a+\epsilon|E, ll_{t-1} = a)$ and substituting in the relation from item #2, we get:

$$\frac{\Pr(ll_t = a + \epsilon|ll_{t-1} = a)}{\Pr(ll_t = a + \epsilon|E, ll_{t-1} = a)} = \frac{e^a}{1+e^a} + \frac{e^{-\epsilon}}{1+e^a}$$

□

Finally, we prove item #4. We make use of a version of item #3 stated in terms of ordinary prices (as opposed to log-likelihood prices):

$$\Pr(p_t = b|E, p_{t-1} = a) = \frac{b}{a}\Pr(p_t = b|p_{t-1} = a)$$

Using this equation and the definition of expectation, we can show that:

$$\begin{aligned} E[p_t|E, p_{t-1} = a] &= \int_0^1 p_t \Pr(p_t|E, p_{t-1} = a)dp_t \\ &= \int_0^1 p_t \cdot \frac{p_t}{a}\Pr(p_t|p_{t-1} = a)dp_t \\ &= \frac{1}{a}\int_0^1 p_t^2 \Pr(p_t|p_{t-1} = a)dp_t \\ &= \frac{1}{a}E[p_t^2|p_{t-1} = a]. \end{aligned}$$

From the definition of variance and the result of item #1, we know that:

$$\begin{aligned} \mathrm{Var}(p_t|p_{t-1} = a) &= E[p_t^2|p_{t-1} = a] - E[p_t|p_{t-1} = a]^2 \\ &= E[p_t^2|p_{t-1} = a] - a^2. \end{aligned}$$

Putting the last two equations together, we have the desired result:

$$E[p_t|E, p_{t-1} = a] = \frac{\mathrm{Var}(p_t|p_{t-1} = a)}{a} + a$$

□

Figure 2 shows the distribution of changes in log-likelihood ($\epsilon = ll_t - ll_{t-1}$) measured across all twenty-two IEM markets.[5] The distribution is nearly symmetric (a positive increase in log-likelihood of $\epsilon$ is almost exactly as likely as a decrease of $\epsilon$), consistent with item #1 of Theorem 1. The plot of Figure 2 aligns almost exactly with the mirror plot of the distribution of $-\epsilon = ll_{t-1} - ll_t$. In fact symmetry is a stronger condition than is provable from Assumption 1 alone; symmetry implies zero expected change (item #1 of Theorem 1) , but not vice versa. Interestingly, the distribution of $\epsilon$ follows a power law over several orders of magnitude. Future work may investigate whether other natural efficiency assumptions can explain the symmetric and power-law behavior of observed distributions.

Figure 3 shows the frequency of changes of $\epsilon$ for candidates that eventually won divided by the frequency of changes of $\epsilon$ for candidates that eventually lost, over a range of $\epsilon$. The solid line plots $e^\epsilon$ as predicted by item #2 of Theorem 1; the fit is reasonably close.

[5]The distribution density is estimated by measuring the cumulative distribution, then approximating the derivative at the $i$th largest $\epsilon$ by $(y_{i-50}-y_{i+50})/(x_{i-50}-x_{i+50})$, where $x_i$ is the value of the $i$th largest $\epsilon$ and $y_i$ is the cumulative distribution at that point.

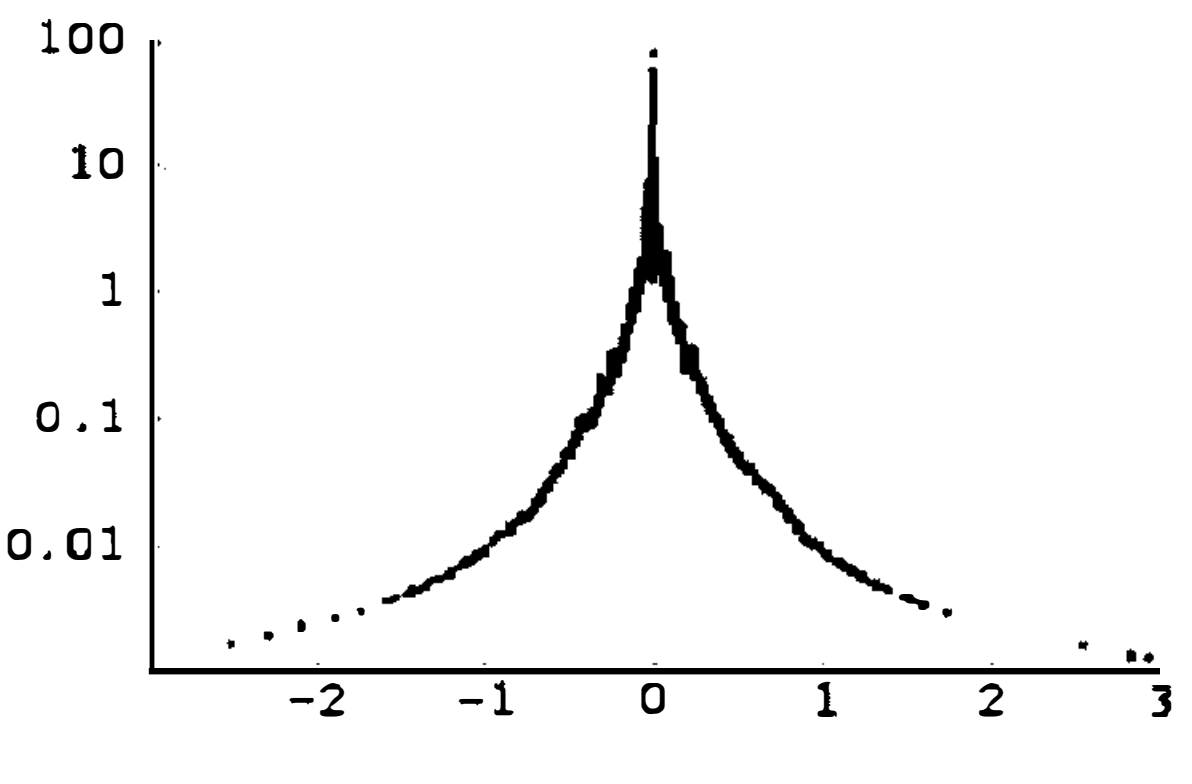

Figure 2: Distribution of changes $\epsilon$ in log-likelihood price measured over twenty-two election markets on IEM.

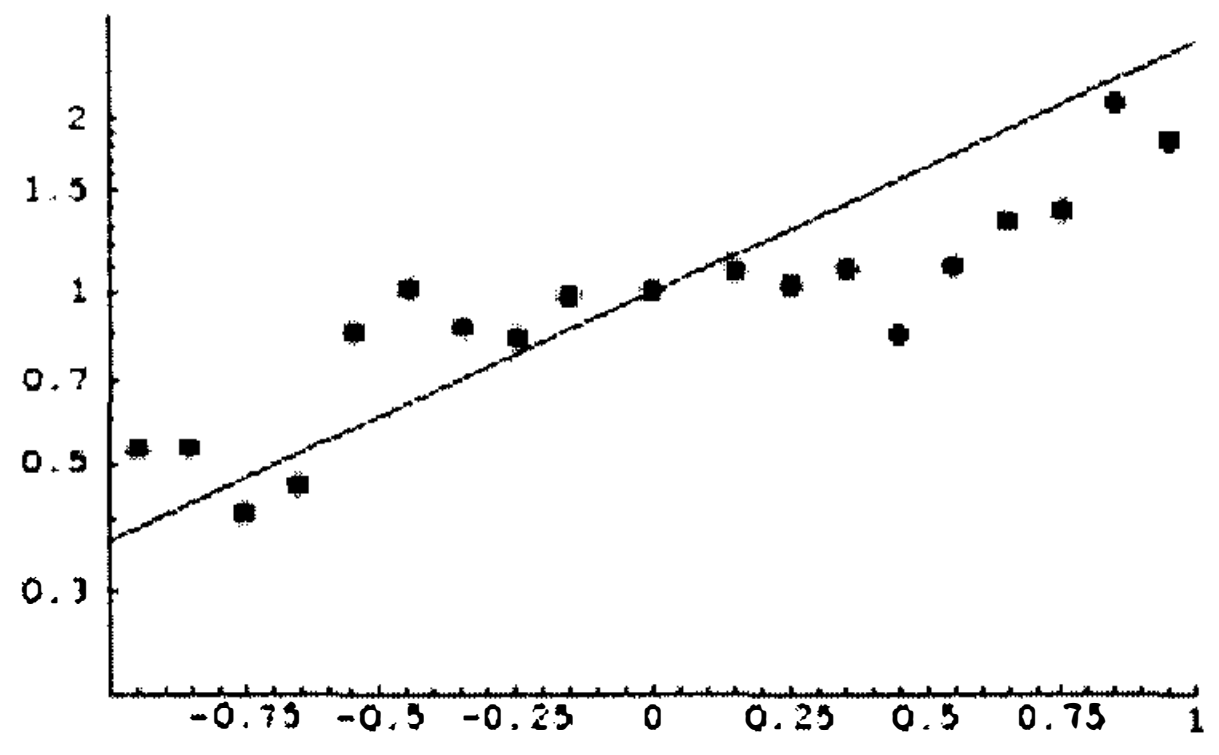

Figure 3: Distribution of changes $\epsilon$ in log-likelihood price for winning candidates divided by changes $\epsilon$ for losing candidates, on a linear-log scale. The line corresponds to $e^\epsilon$, the theoretically predicted relationship.

The fact that the logarithmic score tends to increase over time as seen in Figure 1 can be explained using items #2, #3, and #4 of Theorem 1. According to the definition of logarithmic score, the figure shows the average of the logarithm of the price of the winning candidates. Items #2 and #3 state that the prices of the winning candidates should go up faster on average than the prices of losing candidates, or the prices of all candidates taken together (assuming that there is at least some variation in prices). Item #4 describes the nature of the expected increase most directly. The expected increase in price from one time step to the next (in worlds where $E$ is true) is equal to the variance in price at that time divided by the price. As long as the variance of price is nonzero (i.e., as long as prices have some probability of changing), then there will be an expected increase in price as time moves forward. The following corollary (the proof is immediate from item #4 of Theorem 1) establishes that prices provably trend toward the correct outcome over time.

**Corollary 1** *Assuming some positive variance of future price $p_t$ given the current price $p_{t-1} = a$ (that is, $\mathrm{Var}(p_t|p_{t-1} = a) > 0$), then:*

$$E[p_t|E, p_{t-1} = a] > a.$$

*That is, the expected future price is greater than the current price in worlds where E is true.*

Item #3 of Theorem 1 can also be used to obtain more specific results regarding the average logarithmic score, if more information is known about the apriori distribution of $\epsilon$, for example, if we know that the distribution is a power law.

## 5 MODEL OF INFORMATION INCORPORATION

The previous section showed that several properties of the dynamics of betting markets can be explained with very simple assumptions. The increase in logarithmic score observed on IEM is implied once we assume accuracy and variability of prices; no notion of evidence or information is explicitly needed. In this section, we seek to model information directly, and examine the resulting effect on price dynamics.

We model the event $E$ as the occurrence of $\lceil n/2 \rceil$ or more tails in a series of $n$ fair coin flips. We model the release of information as the revelation of the outcomes of one of more coin flips. At time $t = 0$, the process begins with some apriori knowledge: $i_0$ tails have occurred out of $k_0$ trials. From that point on—from time $t = 1$ to time $t = n - k$—one coin is revealed per time step. At any given time $t$, $i_t$ tails have occurred out of $k_t$ trials in total (including the initial $i_0$ out of $k_0$).

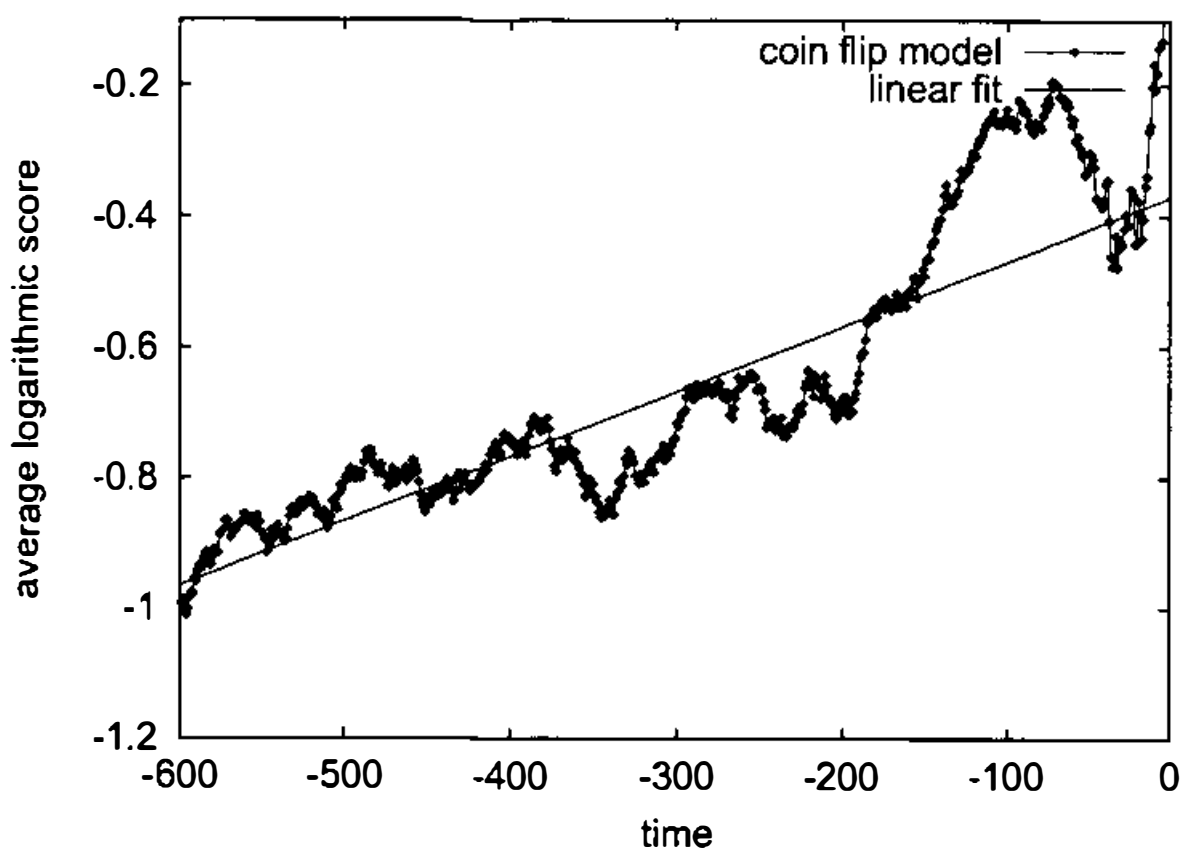

Figure 4: Average logarithmic score of twenty-two simulated markets using the coin flip model.

In order for $E$ to occur, $\lceil n/2 \rceil - i_t$ more tails need to occur in the remaining $n - k_t$ trials. So at time $t$, the probability of the event $E$ occurring is:

$$\Pr(E|i_t, k_t) = \left(\frac{1}{2}\right)^{n-k_t} \sum_{j=\lceil n/2 \rceil - i_t}^{n-k_t} \binom{n-k_t}{j} \quad (1)$$

We again make Assumption 1 that $p_t = \Pr(E|i_t, k_t)$, or, as in RE theory, all information known at time $t$ is incorporated into the price at time $t$. Figure 4 shows the average logarithmic score computed over twenty-two "markets" simulated according to (1) with $i_0 = k_0 = 0$ and $n = 1200$. The figure show the price at every other time step; this corresponds to two coin flips per day when comparing with the IEM graph of Figure 1. We see a similar qualitative pattern of variability, roughly linear increase over a large period of time, and rapid increase near the end. This suggests that the dynamics of prices on IEM can be rationalized as resulting from a process where information is incorporated at roughly a constant rate on average across all markets. There are also discernible differences between Figures 1 and 4: for example, the variability of IEM prices appears greater than in the model.

## 6 DETECTING AND EXPLAINING INFORMATION EVENTS

Previous sections characterized the process of information incorporation in betting markets. In this section, we examine the possibility of using markets to detect significant information events and explain them using alternate textual sources. As a proof of concept, we present a semi-automatic procedure (which we believe is fully automatable) for monitoring markets and extracting explanations from newsgroups and official news outlets. We present results from three markets: two from IEM and one from the Foresight Exchange,[6] a market game that operates like a betting market (much like IEM), except that all transactions are using play money. Our previous studies show that play-money market games behave in many ways like real markets (Pennock, Lawrence, Giles, & Nielsen, 2001a; Pennock, Lawrence, Nielsen, & Giles, 2001b).

The three markets examined were: (1) candidate Giuliani in the 2000 US NY Senate election, (2) candidate Gore in the 2000 US Presidential election, and (3) the outcome "extraterrestrial life discovered by 2050" (XLif) as defined on the Foresight Exchange.[7] Note that in market (2), the winning bet was defined as the candidate with the largest share of the *popular* vote, not the winner of the electoral college, so Gore was the eventual winning bet.

We characterized daily price fluctuations in the three markets using the difference between log-likelihood prices from one day to the next. We identified days on which exceptionally large differences were observed. We found the following dates to be pivotal dates, immediately following huge price swings:

1. April 27, 2000 and May 19, 2000 for candidate Giuliani in NY Senate market
2. November 08, 2000 for candidate Gore in the US Presidential market
3. August 06, 1996 for XLif on the Foresight Exchange

Figures 5, 6, and 7 show the price graphs for the three markets surrounding these dates. We used each of the dates as a splitting point for generating two text corpuses: a *negative* set of documents from before the date in question, and a *positive* set of documents from the week following the date in question. Documents were gathered from Usenet news archives on Google[8] for all three markets. We gathered all postings during relevant date ranges to the newsgroups ny.politics (622 postings), us.politics (480 postings), and sci.space.news (127 postings) for the three markets, respectively. We did not use any keywords to narrow the search further. Additionally, we gathered the titles and abstracts of articles in the Washington Post containing "Giuliani" for the NY Senate market (189 articles).

We identify the features (words and up to three-word phrases) that differentiate the positive and negative document sets using expected entropy loss (Glover, Flake, Lawrence, Birmingham, Kruger, Giles, & Pennock, 2001). We do not explicitly remove stop words. Instead, we remove all features that occur in less than 7.5% of the positive documents, we remove all dates and numbers, and we manually remove source-specific words (e.g., "google" and "Washington Post"). We then rank keywords by expected entropy loss as follows. Entropy is computed independently for each feature. Let $P$ be the event that a document is in the positive set. Let $f$ denote the event that the document contains the specified feature (e.g., contains the word "meteorite"). The prior entropy of the class distribution is $e \equiv -\Pr(P)\lg\Pr(P) - \Pr(\bar{P})\lg\Pr(\bar{P})$. The posterior entropy of the class when the feature is present is $e_f \equiv -\Pr(P|f)\lg\Pr(P|f) - \Pr(\bar{P}|f)\lg\Pr(\bar{P}|f)$; likewise, the posterior entropy of the class when the feature is absent is $e_{\bar{f}} \equiv -\Pr(P|\bar{f})\lg\Pr(P|\bar{f}) - \Pr(\bar{P}|\bar{f})\lg\Pr(\bar{P}|\bar{f})$. Thus, the expected posterior entropy is $e_f\Pr(f) + e_{\bar{f}}\Pr(\bar{f})$, and the *expected entropy loss* is $e - \left(e_f\Pr(f) + e_{\bar{f}}\Pr(\bar{f})\right)$. If any of the probabilities are zero, we use a fixed value instead of 0 in the equations. Expected entropy loss is synonymous with expected information gain, and is *always* nonnegative. Features are sorted by expected entropy loss to provide an approximation of the usefulness of the individual feature. This approach will correctly assign low scores to features that, although common in both sets, are unlikely to be useful for a binary classifier.

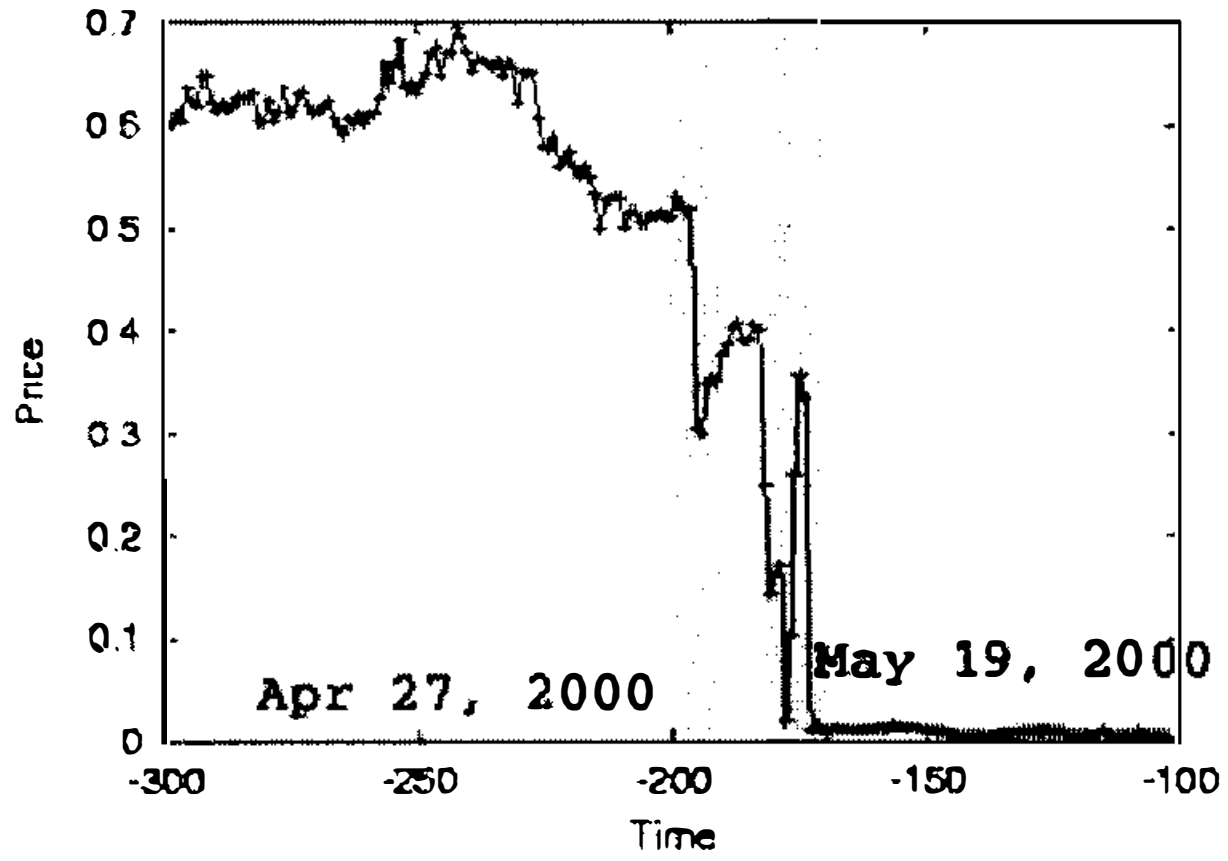

Figure 5: Portion of the price time series for candidate Giuliani in NY Senate election market on IEM, with significant dates highlighted.

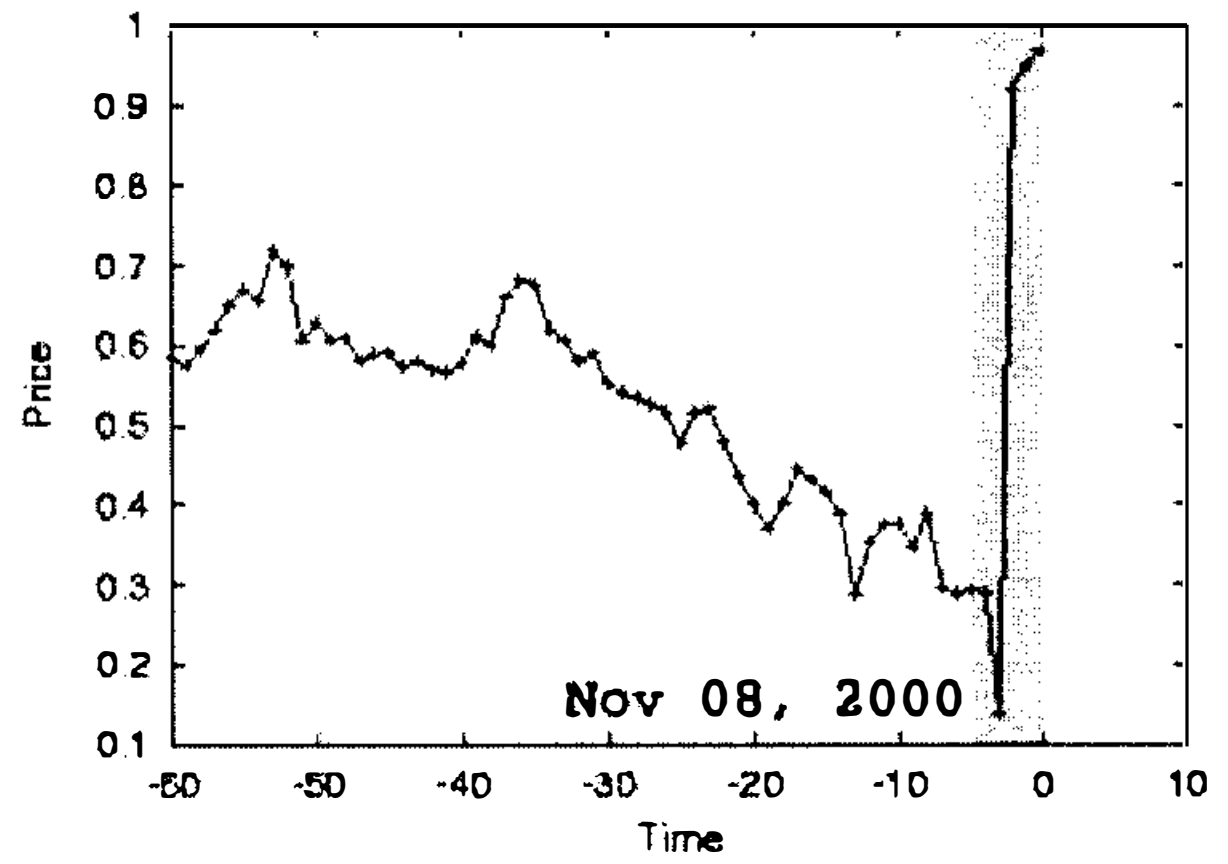

Figure 6: Portion of the price time series for candidate Gore in US Presidential election market on IEM. Note that in this market a bet for Gore won if Gore had a larger share of the *popular* vote.

[6]`http://www.ideosphere.com/`

[7]`http://www.ideosphere.com/fx-bin/Claim?claim=XLif`

[8]`http://groups.google.com/`

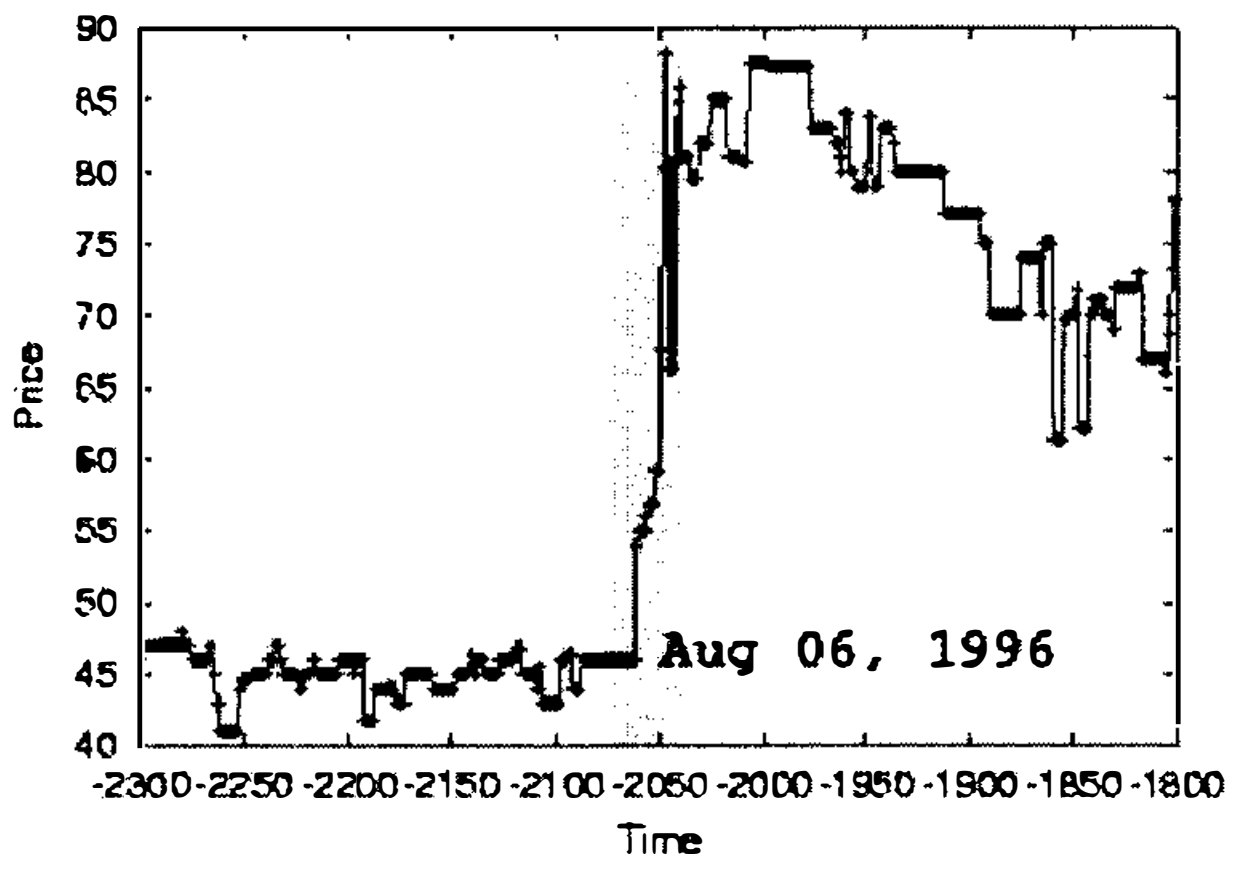

Figure 7: Portion of the price time series for "100 cents if extra-terrestrial life discovered by 2050" on the Foresight Exchange.

| source/date | top ranked features |
|---|---|
| Usenet ny.politics Apr 27, 2000 | cancer, cancer newsgroups,prostate, prostate cancer, cancer newsgroups ny, commies subject liberal, subject liberal propaganda, damnliberals liberals, has prostate, commies |
| Washington Post Apr 27, 2000 | cancer, from prostate, is suffering from, business of politics, diagnosis, suffering from prostate, prostate cancer, suffering, from prostate cancer, cancer diagnosis |
| Usenet ny.politics May 19, 2000 | lazio, rick lazio, mayor, voted, rick, rep rick lazio, alt fan rush, difference, ca politics, families |
| Washington Post May 19, 2000 | lazio, rick lazio, rick, rep rick, rep rick lazio, convinced, opponent, rudy, abortion rights, giuliani's inner |

Table 1: Top features found for dates corresponding to major price changes in the NY Senate market.

Results are shown in the Tables 1, 2, and 3. The procedure did extract many words and phrases which are closely associated with real incidents happening during the identified dates with major implications for the corresponding market bets. For example, April 27, 2000 is the day Giuliani announced he had prostate cancer. Entropy loss extracted terms and phrases like "cancer", "prostate", "prostate cancer" from both Usenet newsgroups and the Washington Post articles. May 19, 2000 is around the time Giuliani formally announced he was quitting the Senate race, with Rick Lazio the replacement Republican candidate. Again reasonable explanatory terms and phrases were discovered using our algorithm. Both the Usenet results ("lazio", "rick lazio", "mayor", "voted" ) and Washington Post results ("lazio", "rick lazio", "rick", "rep rick", "rep rick lazio") indicate the name of Lazio as a top ranked feature. Words like "drop", "dropped", "quit", and "bow out" did appear in the positive documents, but were removed during thresholding. We believe that more intelligent use of stemming and synonyms would help in this situation where there are many ways to say the same thing (as opposed to the case of "prostate cancer", where there is essentially only one way to say it).

In the US Presidential election, the price of candidate Gore skyrocketed after the election when it became clear he won the popular vote. The near tie and resulting chaos in counting the ballots in Florida appear in our extracted keyword list, where the top ranked features are "florida", "ballots", "recount", "palm beach" etc.

On August 6, 1996, NASA announced it had discovered possible signs of life on a Martian meteorite. The price of a bet on XLif on the Foresight Exchange rose quickly, apparently in response. Indeed our algorithm found very relevant explanatory features, as listed in Table 3.

| source/date | top ranked features |
|---|---|
| Usenet us.politics Nov 08, 2000 | florida, ballots, recount, palm beach, ballot, beach county, palm beach county, recounts, counted, county, in palm, the ballot, counties, fraud, in palm beach |

Table 2: Top features found for the US Presidential market.

| source/date | top ranked features |
|---|---|
| Usenet sci.space.news Aug 06, 1996 | meteorite, life, evidence, washington dc august, martian meteorite, primitive, gibson, organic, of possible, martian, life on mars, david, life on, billion years ago, mckay |

Table 3: Top features found for the XLif market on the Foresight Exchange.

## 7 CONCLUSION

In this paper, we explored the mechanism of information incorporation in betting markets. We developed a theory based on simple efficiency assumptions that explains the increase in forecast accuracy over time observed in real betting markets. We proposed a simple coin-flipping model of information flow into a market and demonstrated a qualitative correspondence with real data. We designed an algorithm for detecting information events in real markets and explaining them by extracting features from online text sources. In three case studies, our algorithm found key words and phrases subjectively very relevant to the events of the day corresponding to sharp market upswings or downswings.

### Acknowledgments

We thank Finn Nielsen and Steve Lawrence for helpful ideas and insights.